\documentclass[letterpaper]{article} 
\usepackage{aaai2026}  
\usepackage{times}  
\usepackage{helvet}  
\usepackage{courier}  
\usepackage[hyphens]{url}  
\usepackage{graphicx} 
\usepackage{natbib}  
\usepackage{caption} 
\usepackage{algorithm}
\usepackage{algorithmic}
\usepackage{makecell}
\usepackage{multirow}
\usepackage[most]{tcolorbox}

\usepackage{colortbl}

\usepackage{newfloat}
\usepackage{listings}
\DeclareCaptionStyle{ruled}{labelfont=normalfont,labelsep=colon,strut=off} 
\floatstyle{ruled}
\newfloat{listing}{tb}{lst}{}
\floatname{listing}{Listing}
 \nocopyright

\title{Doing Less with More: A First-Principles Exploration of the Suitability of Agentic Computing over Alternative Architectural Choices}

\author {
    Ritvik Garimella\textsuperscript{\rm 1},
    Biplav Srivastava\textsuperscript{\rm 1},
    Amit Sheth\textsuperscript{\rm 1,\rm 2}
}
\affiliations {
    \textsuperscript{\rm 1}Artificial Intelligence Institute, University of South Carolina\\
    \textsuperscript{\rm 2}Indian AI Research Organization\\
    \textbf{Correspondence: }ritvikg@sc.edu
}

\usepackage{xcolor}

\begin{document}

\maketitle

\begin{abstract}

There is growing interest in automating business activities with Agentic Artificial Intelligence (AI) due to latter's seeming ease of use. \textit{Never has it been easier, or costlier, to do less with more.} However, little is known about when agents are preferable to established alternatives such as local computation, Representational State Transfer (REST), the Simple Object Access Protocol (SOAP), and the Model Context Protocol (MCP), particularly when development speed, performance, and operational cost are considered. We investigate this question using a controlled mathematical task that compares seven methods on a benchmark of 1{,}000 arithmetic expressions where semantics of operator precedence has to be preserved. We ran this setup across a native Function Calling baseline (Python library), REST, SOAP and MCP microservice architectures, direct responses from large language models (LLMs) using a Groq cloud API and five locally deployed Ollama models, and agentic systems in which five local LLMs invoke MCP tools. The deterministic methods achieve 100\% accuracy with negligible latency. Direct LLM evaluation achieves at most 92.9\% accuracy at over 22{,}000$\times$ the latency of a native function call. With MCP tools, the best agent reaches 99.9\% accuracy, but incurs nearly 1{,}000{,}000$\times$ the baseline latency. These results demonstrate that LLMs approaches are poorly suited for exact arithmetic evaluation and motivate an informed \textbf{architectural decision framework (ADF)} for choosing an architecture based on accuracy and throughput requirements. We recognize that an actual interaction with an agent may span both reasoning and linguistic tasks. Hence our recommendation is to explore multiple architectural choices (based on task type) rather than one-size-fits-all solution (LLMs).

Code is available at - \url{https://github.com/Ritvik-G/adf/}
\end{abstract}


\section{Introduction}
\label{sec:Introduction}

The prevalent approach of building large software systems is to modularize code as functions and run them locally or remotely \citep{modularization}. The mechanisms to implement code modularity, however, has evolved over time with technologies ranging from local function call, Representational State Transfer (REST), Simple Object Access Protocol (SOAP), the Model Context Protocol (MCP)~\citep{mcp2024} to calls now orchestrated and executed using Large Language Models (LLMs) \cite{API, restful, rodriguez2008restful, weerawarana2005web, halili2018web, schick2023toolformer, yao2023react}. Each generation of this evolution has brought both opportunities and trade-offs with respect to development speed, runtime performance, and operating cost.
However, surprisingly, these trade-off aspects are rarely analyzed and quantified under similar tasks and conditions.

Even when a comparison may be done, 
care is needed as simplistic comparisons  risk conflating the computation's outcome (functional effectiveness like accuracy) itself with the operational performance of mechanism used to invoke it. For example, the choice between
REST and SOAP or MCP typically does not expose an underlying deterministic implementation, whereas an LLM asked to produce an answer replaces exact execution with probabilistic inference.
Furthermore, an orthogonal performance issue with code modularity is when to call the functions leveraging potential parallelism leading to various strategies for composition, orchestration and execution \citep{biplav-web-services-composition-execution}. 

In this paper, we do a comparison of architectural choices for functional invocation in the context of arithmetic expression evaluation. Here, the semantics of operator precedence plays a direct role in accuracy of computation while the choice of architecture impacts performance of (chosen workflow's) composition and execution. Illustrated in the example below (bracket added for readability only):

\begin{tcolorbox}[colback=gray!5,colframe=gray!50!black,
title=Example Arithmetic Expression (from dataset)]
\textbf{Expression:} $(78.0 * ( 186.0 - 172.0 )) $
\\
\textbf{Expected Answer:} $1092$
\end{tcolorbox}
Consider the semantics of the above situation, the operator precedence is crucial. If multiplication was done first, it would result in an incorrect answer of 14336. Due to automatic verifiability of the answer and dependency on semantics of operator precedence (in this case, BODMAS - also called  PEMDAS rules), this task plays a useful control setting in isolating the reasoning problem. Our goal from this work is not to argue that arithmetic expression evaluation as a representation for reasoning tasks in general, but to use it as a primer for a broader point i.e., every invocation mechanism has a place, there is no one-size fits all scenario (LLMs) and choices should be made with consideration of tradeoffs (described more in Section \ref{sec:framework}).


We compare seven representative strategies for implementing functionality (mathematical operations in our case) spanning four architectural families: (i) direct function calling; (ii) service-oriented architectures (SOA) through REST, SOAP and MCP; (iii) LLMs answering directly, hosted locally and remotely; and (iv) agentic LLMs that call the arithmetic operations as MCP tools. All strategies are evaluated on the same set of 1{,}000 arithmetic expressions extracted from SVAMP~\citep{svamp}, measuring exact-answer accuracy, execution time and processing cost (Section~\ref{sec:methodology}). Our results show that deterministic methods (Function Calling, REST, SOAP and MCP) are both the most accurate and the fastest. LLMs answering directly trail substantially on both measures, varying widely by model and remaining insensitive to temperature. Giving LLMs deterministic tools recovers accuracy only for models that reliably use them, at a further cost in latency. Direct LLM calls were the easiest to assemble (in lines of code), but this ease of integration does not guarantee the correctness that exact arithmetic demands (Section~\ref{sec:results}).

\begin{tcolorbox}[colback=gray!5,colframe=gray!50!black,
title=Key Contributions]
\begin{itemize}
    \item A controlled evaluation of leading alternative choices for algebraic evaluation exploring accuracy and performance.
    \item An architectural decision framework for selecting an invocation strategy according to tradeoff considerations. 
\end{itemize}
\end{tcolorbox}

The remainder of the paper is organized as follows. Section~\ref{sec:background} introduces the relevant computation and invocation paradigms. Section~\ref{sec:methodology} describes the experimental design, benchmark construction, evaluated configurations, and metrics. Section~\ref{sec:results} presents the empirical results. Section~\ref{sec:framework} --~\ref{sec:discussion} discusses their architectural implications and the resulting decision framework. Section~\ref{sec:conclusion} concludes the paper.

\begin{figure*}[t]
  \centering
  \includegraphics[width=1\textwidth]{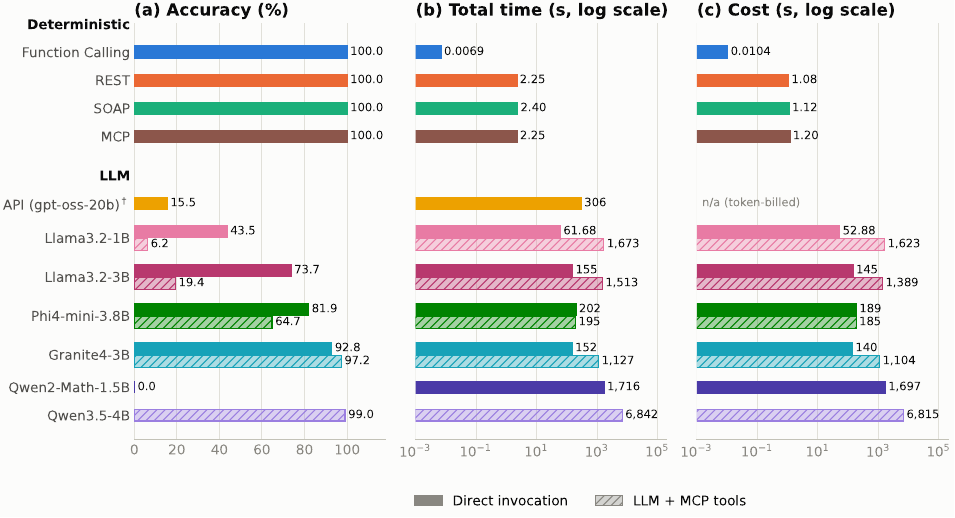}
  \caption{\textbf{Performance comparison across evaluation measures.} (a) Accuracy (higher is better); (b) total execution time and (c) cost, both in seconds on a logarithmic scale (lower is better). Cost is client CPU time for deterministic methods and Ollama-reported inference time for local LLMs. Solid bars show direct invocation and hatched bars the same model with MCP tools; LLM values are averaged over $T=0$ and $T=0.7$.}
  \label{fig:measures}
\end{figure*}

\begin{figure*}[h]
  \centering
  \includegraphics[width=0.9\textwidth]{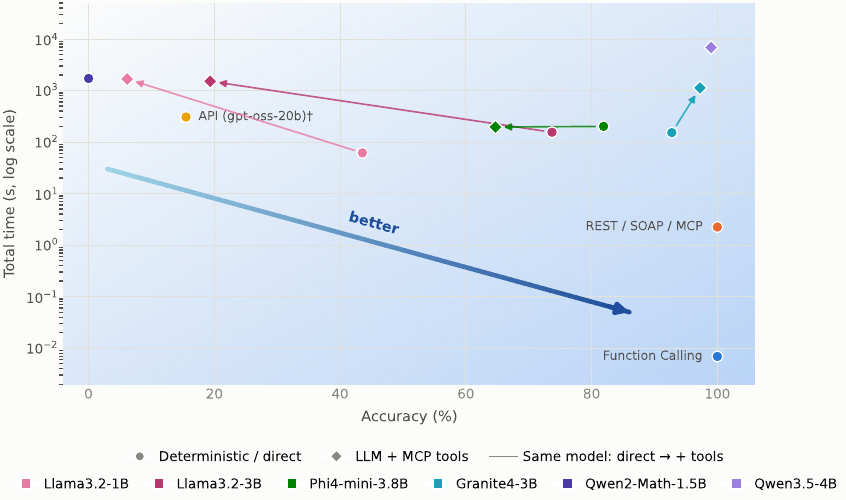}
  \caption{\textbf{Accuracy–efficiency trade-off across all configurations.} Accuracy against total execution time (logarithmic scale), averaged over $T=0$ and $T=0.7$. Circles denote deterministic methods and direct LLM invocation, diamonds LLMs with MCP tools; arrows connect each model's direct and with-tools results. REST, SOAP and MCP overlap. \textbf{Bottom right is better.}}
  \label{fig:efficiency}
\end{figure*}

\begin{figure*}[t]
  \centering
  \includegraphics[width=1\textwidth]{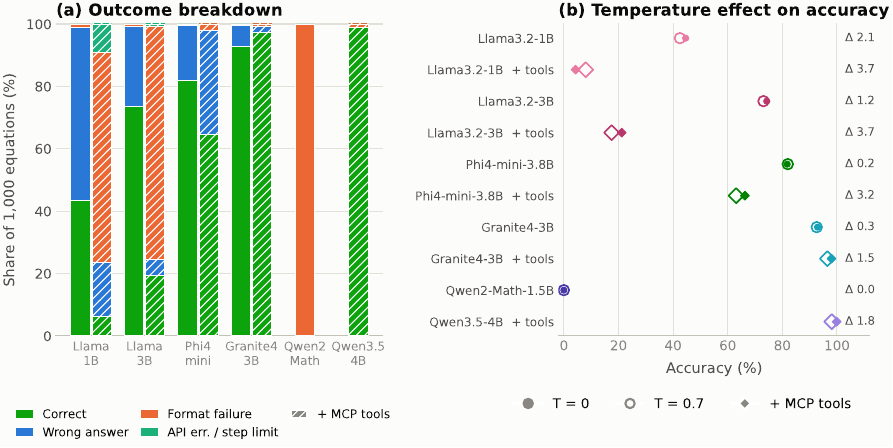}
  \caption{\textbf{Outcome distribution and Effect of temperature on model accuracy and outcome distribution. }(a) Outcome breakdown of local LLMs, answering directly (solid) and with MCP tools (hatched), averaged over $T=0$ and $T=0.7$; API error/step limit covers rejected requests and agents exceeding six turns. (b) Accuracy at $T=0$ (filled) and $T=0.7$ (hollow) for every model and mode; $\Delta$ gives the difference, at most 3.7 points.}
  \label{fig:breakdown}
\end{figure*}

\section{Background}
\label{sec:background}

Agentic AI is widely used to automate tasks that could equally be handled by simpler, deterministic mechanisms ~\citep{schick2023toolformer,yao2023react}. Before going directly into one-size-fits-all approach, it is worth asking to what extent can they actually cause an impact and whether local existing mechanisms (local function call, REST endpoint, SOAP service, MCP server) can serve as a better solution. This section briefly reviews these mechanisms and the LLM-based alternatives (direct and agentic), the role of execution order in composing them and why exact computation is not guaranteed via LLM's decoding process. 

\textbf{Invocation Paradigms.} In-process function call transfer control directly to locally available implementation with no overheads. This makes them a natural baseline whenever an implementation procedure is well known (e.g., PEMDAS rules of arithmetic expressions). 

REST and SOAP expose the same kind of computation across a process and network boundary over HTTP ~\citep{restful,rodriguez2008restful,weerawarana2005web}. REST uses lightweight, resource-oriented endpoints whereas SOAP wraps calls in XML envelopes validated against an explicit contract (WSDL).  Because both approaches share the same transport, their overhead should be comparable with SOAP's envelope and schema validation being a more structured definition \citep{castillo2011soaprest,potts2018soaprest}. SOAP also supports dynamic service discovery, which we do not evaluate.




The Model Context Protocol (MCP)~\citep{mcp2024} standardizes how applications expose tools to clients. An MCP server lists each tool with a name, a description and a JSON schema for its arguments, and clients invoke tools through JSON-RPC calls over HTTP. When called from deterministic code, MCP is simply another remote-invocation protocol. Its tools-calling is optimized for LLM-based invocations where an LLM must discover and select tools at runtime, a role similar to SOAP's service discovery. In our setup, REST, SOAP and MCP all decompose an expression into a sequence of pairwise operations and dispatch them one call at a time.



LLM-based computation replaces this pipeline with a single text generation. Instead of applying a fixed procedure, the model receives a prompt that asks for the numerical answer alone, and correctness depends on its ability to reason about operator precedence. Unlike an interpreter, this offers no guarantee of correctness, since the answer is generated token by token rather than derived~\citep{boye2025mathfailures,shrestha2025gsmranges}. We refer to this direct answer generation as \emph{LLM-based computation}, and to the setting in which the LLM instead selects and calls deterministic tools (here, MCP tools) and only composes their results as \emph{agentic computation}~\citep{schick2023toolformer,yao2023react}.

\textbf{Order-Preserving Computation.} Arithmetic expressions make correctness easy to check but not trivial to preserve. A deterministic evaluator enforces the operator precedence (PEMDAS/BODMAS) through a parse tree. This requires the LLM to understand the semantics of the operator precedence implicitly from the prompt text making it a non-trivial task.

\textbf{Decoding and Failure Modes.} Decoding strategy is assumed to be a highly influential variable in LLMs with Temperature ($T$) used as the hyperparameter. Greedy decoding ($T=0$) always picks the highest-probability token whereas most LLMs use a temperature $T=0.7$ to introduce stochasticity (allowing for exploration and ideal for text generation) ~\citep{renze2024temperature}. Beyond wrong answers, an LLM can fail by format (a response that is not a number alone), and an agent can additionally fail by ignoring its tools, misusing them, or never reaching a final answer.

\section{Experimental Setup}
\label{sec:methodology}


In this section, we look into the setup used to compare the invocation strategies. We first describe the dataset and ground-truth extraction mechanism, then the implementation details of each method, followed by the metrics used to evaluate accuracy, time and cost. Finally, we talk about the execution conditions under which the methods were run to ensure a fair comparison.

\textbf{Dataset Preparation.} The SVAMP benchmark dataset \cite{svamp}  consists of 1{,}000 mathematical equation strings covering arithmetic operations of varying complexity. Equations were drawn from a combined train/test split and evaluated under identical conditions for all methods.Ground-truth labels are the exact numerical results with input being only the 'equation' part of the dataset (Example: $equation = ( 290.0 / 2.0 )$, $ground\_truth = 145$).

\subsection{Implementation Details}
Each model receives an equation string and must return the correct numerical result. We define seven methods as follows:

\begin{itemize}
    \item \textbf{Function Calling:} Python's built-in \texttt{eval()} function, executing entirely in-process with no network overhead.
    \item \textbf{REST:} Equations are decomposed into sub-operations dispatched as HTTP POST requests to a local Flask server (\texttt{localhost:5001}).
    \item \textbf{SOAP:} We run both the functional services and discovery on the same server using XML-encoded SOAP envelopes over HTTP (\texttt{localhost:8000}). Same decomposition strategy as REST is used.
    \item \textbf{MCP:} Same decomposition as REST, with each sub-operation sent as an MCP tool call (JSON-RPC over HTTP) to a server exposing \texttt{add}, \texttt{subtract}, \texttt{multiply} and \texttt{divide} as tools (\texttt{localhost:8100}).
    \item \textbf{LLM-API:} Each equation is submitted as a natural language prompt to the Groq cloud API (\texttt{openai/gpt-oss-20b}, $\sim$20B parameters), instructed to return only a numerical answer.
    \item \textbf{Ollama LLM:} Identical prompting strategy is executed over five locally hosted models \texttt{llama3.2:1b, llama3.2:3b, phi4-mini:3.8b, granite4:3b, qwen2-math:1.5b} via the Ollama runtime (\texttt{localhost:11434}). 
    \item \textbf{LLM + MCP Tools:} Local models receive the MCP server's four tools as function definitions and must solve each equation by calling them. Each tool call is executed on the MCP server and its result returned to the model, until the model replies without a tool call (its final answer) or six LLM turns pass. Each reply is limited to 256 tokens.. Models: \texttt{llama3.2:1b, llama3.2:3b, phi4-mini:3.8b,  granite4:3b, qwen3.5:4b}. Qwen2-Math-1.5B is excluded as it does not support tool calling, instead substituted with Qwen3.5-4B.
\end{itemize}

All LLM calls use a sampling temperature of 0.7, a common temperature setting for NLU tasks. They were once again run with temperature of 0 for greedy decoded response. Responses were parsed to only select numeric token; unparseable responses were counted as incorrect.

\subsection{Evaluation Principles}
Each method was assessed on four dimensions: Accuracy, Time, Cost and Development Effort. Each were measured as follows:
\begin{itemize}
\item \textbf{Accuracy}: The percentage of exactly correct numerical output from the invocations. For LLMs, this meant that any non-numeric or explanatory text in the response (e.g., "The answer is..." ) was treated as a failure mode and bucketed under format failure, resulting in the response marked as incorrect answer. 
\item \textbf{Time}: Total wall clock time from the start of the program to the end. This was then divided by number of samples (1000) for each method to get avg. time per sample.
\item \textbf{Cost}: The cost of the methods were calculated using CPU-seconds for deterministic pipelines such as Function Calling, REST, SOAP and MCP (single delta using \texttt{psutil}'s process-level counters). The local LLMs use ollama-reported inference duration (prompt evaluation + generation time), reflecting both CPU and GPU execution costs. These are the total amount of processor time a process consumed during inference collectively called Runtime Cost and measured in seconds. For cloud API, we report token consumption instead. For LLM + MCP Tools, this is summed over every LLM turn.

\item \textbf{Development Effort}: We measure the effort taken to develop software by lines-of-code. Although there are many measures in software engineering literature for effort, code length is easy to calculate and relate to across architecture styles.

\end{itemize}

\subsection{Execution Strategy}
All experiments were executed on the same host machine; Cloud inference occurred remotely on Groq hardware (with fixed temperature of 0.7), while local Ollama inferences ran with two temperature settings ($T=0.7$ and $T=0$) to emphasize both general chatbot level behaviour and greedy decoding, which is often mistook as "determinism" choice by many \citep{deterministic}. For LLM based invocations (API and Ollama), a single prompt was used as a primer while agentic (LLM + MCP Tools) used the second as shown below: 
\begin{tcolorbox}[colback=gray!5,colframe=gray!50!black,
title=LLM Prompt]
Answer with ONLY the final numerical value, no explanations or equations. \\
\textbf{Question:} $\{equation\} $
\\
\textbf{Answer:}
\end{tcolorbox}

\begin{tcolorbox}[colback=gray!5,colframe=gray!50!black,
title=LLM + MCP Tools Prompt]
\textbf{System:} You are a calculator agent. Use the provided tools for every arithmetic operation, one operation per tool call. When you have the final result, reply with ONLY the final numerical value, no explanations or equations. \\
\textbf{User:} QUESTION: $\{equation\}$ ANSWER:
\end{tcolorbox}

\section{Results}
\label{sec:results}

\begin{table*}[t]
\centering
\begin{tabular}{lccccccc}
\hline
\small
\textbf{Method}  & \textbf{Acc. (\%)} & \textbf{T.Time(s)} & \textbf{Avg.Time/Smpl(ms)} & \textbf{Time$\times$} & \textbf{Process Cost(s)} & \textbf{Cost$\times$} \\
\hline
\rowcolor{gray!15} \multicolumn{7}{l}{\textbf{\textit{Type-1: Deterministic}}} \\
Function Calling* & 100\% & \textbf{0.0069} & \textbf{0.007} & 1$\times$ & \textbf{0.0104} & 1$\times$ \\
REST & 100\% & 2.25 & 2.25 & 326$\times$ & 1.0782 & 104$\times$ \\
SOAP & 100\% & 2.40 & 2.40 & 347$\times$ & 1.1219 & 108$\times$ \\
MCP & 100\% & 2.25 & 2.25 & 326$\times$ & 1.2016 & 116$\times$ \\
\hline
\rowcolor{gray!15} \multicolumn{7}{l}{\textit{\textbf{Type-2: Direct invocation (LLM)}}} \\
API (gpt-oss-20b)$^\dagger$ & 15.5\% & 306.35 & 306.35 & 44{,}399$\times$ & n/a & n/a \\
Llama3.2-1B (T=0) & 44.6\% & 62.64 & 62.64 & 9{,}078$\times$ & 53.12 & 5{,}108$\times$ \\
Llama3.2-1B (T=0.7) & 42.5\% & 60.72 & 60.72 & 8{,}800$\times$ & 52.65 & 5{,}063$\times$ \\
Llama3.2-3B (T=0) & 74.3\% & 157.27 & 157.27 & 22{,}793$\times$ & 146.19 & 14{,}057$\times$ \\
Llama3.2-3B (T=0.7) & 73.1\% & 153.04 & 153.04 & 22{,}179$\times$ & 144.06 & 13{,}852$\times$ \\
Phi4-mini-3.8B (T=0) & 81.8\% & 205.29 & 205.29 & 29{,}752$\times$ & 190.95 & 18{,}360$\times$ \\
Phi4-mini-3.8B (T=0.7) & 82.0\% & 198.00 & 198.00 & 28{,}695$\times$ & 187.58 & 18{,}037$\times$ \\
Granite4-3B (T=0) & 92.9\% & 153.51 & 153.51 & 22{,}248$\times$ & 140.27 & 13{,}488$\times$ \\
Granite4-3B (T=0.7) & 92.6\% & 151.31 & 151.31 & 21{,}929$\times$ & 140.22 & 13{,}483$\times$ \\
Qwen2-Math-1.5B (T=0) & 0.0\% & 1658.21 & 1658.21 & 240{,}320$\times$ & 1643.48 & 158{,}027$\times$ \\
Qwen2-Math-1.5B (T=0.7) & 0.0\% & 1772.81 & 1772.81 & 256{,}929$\times$ & 1751.03 & 168{,}368$\times$ \\
\hline
\rowcolor{gray!15} \multicolumn{7}{l}{\textbf{\textit{Type-3: LLM + Tools (MCP)}}} \\
Llama3.2-1B (T=0) & 4.3\% & 1796.73 & 1796.73 & 260{,}395$\times$ & 1757.02 & 168{,}945$\times$ \\
Llama3.2-1B (T=0.7) & 8.0\% & 1550.03 & 1550.03 & 224{,}642$\times$ & 1488.30 & 143{,}106$\times$ \\
Llama3.2-3B (T=0) & 21.2\% & 1509.96 & 1509.96 & 218{,}835$\times$ & 1402.50 & 134{,}856$\times$ \\
Llama3.2-3B (T=0.7) & 17.5\% & 1515.96 & 1515.96 & 219{,}705$\times$ & 1376.06 & 132{,}314$\times$ \\
Phi4-mini-3.8B (T=0) & 66.3\% & 197.09 & 197.09 & 28{,}563$\times$ & 185.05 & 17{,}794$\times$ \\
Phi4-mini-3.8B (T=0.7) & 63.1\% & 193.56 & 193.56 & 28{,}053$\times$ & 184.36 & 17{,}727$\times$ \\
Granite4-3B (T=0) & 98.0\% & 1126.79 & 1126.79 & 163{,}303$\times$ & 1103.59 & 106{,}114$\times$ \\
Granite4-3B (T=0.7) & 96.5\% & 1126.85 & 1126.85 & 163{,}311$\times$ & 1105.32 & 106{,}281$\times$ \\
Qwen3.5-4B (T=0) & 99.9\% & 6834.43 & 6834.43 & 990{,}497$\times$ & 6805.75 & 654{,}399$\times$ \\
Qwen3.5-4B (T=0.7) & 98.1\% & 6849.51 & 6849.51 & 992{,}683$\times$ & 6825.10 & 656{,}259$\times$ \\
\hline
\end{tabular}
\caption{Benchmark results across all evaluation methods on 1{,}000 equations. Times $\geq$1s are shown to two decimal places. \textbf{Time$\times$} and \textbf{Cost$\times$} report the slowdown/overhead factor relative to the Function Calling baseline (Total Time and Processor Runtime columns, respectively). For local Ollama models, Processor Runtime is Ollama's own self-reported inference compute time (prompt processing + generation duration) and reflects both CPU and GPU execution; for LLM + Tools it is summed over every LLM turn of an equation. \textbf{LLM + Tools} gives each model four MCP tools (add, subtract, multiply, divide) to call instead of computing the answer itself. Qwen2-Math-1.5B does not support tool calling and appears only under direct invocation; Qwen3.5-4B appears only with tools, as answering directly it frequently exceeded the 30\,s request timeout. API invocation cost 39{,}180 tokens in total (15{,}571 prompt + 23{,}609 completion). \textit{*Function Calling is the baseline against which the other methods are compared.}}
\label{tab:results}
\end{table*}

\begin{table}[h]
\centering
\small
\begin{tabular}{lccc}
\hline
\textbf{Model} & \textbf{Correct} & \textbf{Wrong} & \textbf{Format} \\
\hline
\rowcolor{gray!15} \multicolumn{4}{l}{\textbf{\textit{Type-2: Direct invocation (LLM)}}} \\
Llama3.2-1B (T=0) & 446 & 540 & 14 \\
Llama3.2-1B (T=0.7) & 425 & 567 & 8 \\
Llama3.2-3B (T=0) & 743 & 249 & 8 \\
Llama3.2-3B (T=0.7) & 731 & 261 & 8 \\
Phi4-mini-3.8B (T=0) & 818 & 177 & 5 \\
Phi4-mini-3.8B (T=0.7) & 820 & 178 & 2 \\
Granite4-3B (T=0) & 929 & 66 & 5 \\
Granite4-3B (T=0.7) & 926 & 69 & 5 \\
Qwen2-Math-1.5B (T=0) & 0 & 0 & 1000 \\
Qwen2-Math-1.5B (T=0.7) & 0 & 0 & 1000 \\
\hline
\rowcolor{gray!15} \multicolumn{4}{l}{\textbf{\textit{Type-3: LLM + Tools (MCP)}}} \\
Llama3.2-1B (T=0) & 43 & 160 & 700 \\
Llama3.2-1B (T=0.7) & 80 & 191 & 641 \\
Llama3.2-3B (T=0) & 212 & 45 & 740 \\
Llama3.2-3B (T=0.7) & 175 & 59 & 755 \\
Phi4-mini-3.8B (T=0) & 663 & 318 & 19 \\
Phi4-mini-3.8B (T=0.7) & 631 & 347 & 22 \\
Granite4-3B (T=0) & 980 & 17 & 3 \\
Granite4-3B (T=0.7) & 965 & 22 & 13 \\
Qwen3.5-4B (T=0) & 999 & 0 & 1 \\
Qwen3.5-4B (T=0.7) & 981 & 0 & 19 \\
\hline
\end{tabular}
\caption{Outcome breakdown for local Ollama models (n=1{,}000 per configuration). Wrong answers are responses with wrong numbers, whereas format failures are responses that are not a number alone. For LLM + Tools, any remaining equations exceeded the six-turn limit (Section~\ref{sec:results}).}
\label{tab:outcomes}
\end{table}

Table~\ref{tab:results} reports accuracy, time and cost for all 25 evaluated configurations, and Table~\ref{tab:outcomes} decomposes the failure modes of the local LLM executions.

\textbf{Accuracy. }Function Calling, REST, SOAP and MCP achieve 100\% accuracy through deterministic computation. Every LLM-based method fell short of this ceiling, and by widely varying margins. Granite4-3B is the strongest at 92.6--92.9\%, followed by Phi4-mini-3.8B (81.8--82.0\%), Llama3.2-3B (73.1--74.3\%) and Llama3.2-1B (42.5--44.6\%). Qwen2-Math-1.5B scores 0.0\% at both temperatures despite being fine-tuned for math, which reflects a failure to follow the output instruction rather than to compute (Table~\ref{tab:outcomes}). The remote API (gpt-oss-20b) scores 15.5\%.

\textbf{Time. }Wall-clock time spans six orders of magnitude, so Figures~\ref{fig:measures} and~\ref{fig:efficiency} use a logarithmic scale. Function Calling completes all 1{,}000 equations in 6.9\,ms (7\,\textmu s per equation). REST, MCP and SOAP are 326$\times$, 326$\times$ and 347$\times$ slower (2.25--2.40\,s in total), mostly due to HTTP request and response serialization. LLMs answering directly are another one to three orders of magnitude slower. Relative to the baseline, Llama3.2-1B is 8{,}800--9{,}078$\times$ slower (61--63\,ms per equation), Granite4-3B and Llama3.2-3B about 22{,}000$\times$ (151--157\,ms), Phi4-mini-3.8B 28{,}695--29{,}752$\times$ (198--205\,ms) and the remote API 44{,}399$\times$ (306\,ms). Qwen2-Math-1.5B is the slowest at 240{,}320--256{,}929$\times$ (1.66--1.77\,s), as it writes out full step-by-step solutions. Tool use adds up to another order of magnitude, since agents take two to three LLM turns per equation. Granite4-3B slows to 163{,}303--163{,}311$\times$ (1.13\,s per equation) and Qwen3.5-4B reaches 990{,}497--992{,}683$\times$ (6.8\,s). Phi4-mini-3.8B is unchanged at 28{,}053--28{,}563$\times$ because it never calls a tool.

\textbf{Cost. }Cost follows the same pattern. Function Calling uses 0.0104\,s of CPU time, and REST, SOAP and MCP use 1.08--1.20\,s (104--116$\times$). For the cloud API we report tokens instead, 39{,}180 in total (15{,}571 prompt and 23{,}609 completion). Among local models answering directly, Llama3.2-1B uses 52.65--53.12\,s of inference time (5{,}063--5{,}108$\times$), Granite4-3B and Llama3.2-3B 140--146\,s (13{,}483--14{,}057$\times$), Phi4-mini-3.8B 187.58--190.95\,s (18{,}037--18{,}360$\times$) and Qwen2-Math-1.5B 1{,}643--1{,}751\,s (158{,}027--168{,}368$\times$), driven by its long outputs. With tools, cost rises by one to two orders of magnitude for the models that call them. Granite4-3B grows from 140\,s to 1{,}104--1{,}105\,s (106{,}114--106{,}281$\times$), Llama3.2-1B from 53\,s to 1{,}488--1{,}757\,s, and Qwen3.5-4B reaches 6{,}806--6{,}825\,s (654{,}399--656{,}259$\times$).

\textbf{LLM + Tools. }Tools help only the models that use them reliably. Qwen3.5-4B reaches 98.1--99.9\%, and Granite4-3B improves from 92.6--92.9\% to 96.5--98.0\%. Qwen3.5-4B makes 1.22--1.24 tool calls per equation, matching the dataset's 1.236 operations per equation, and Granite4-3B makes 1.19. The other models get worse. Phi4-mini-3.8B never calls a tool and drops to 63.1--66.3\%, below its direct accuracy. The Llama models call tools (1.44--1.73 per equation) but misuse them, with 110--189 tool errors per 1{,}000 equations, and then mostly answer in prose, falling to 17.5--21.2\% (3B) and 4.3--8.0\% (1B). Llama3.2-1B also exceeds the six-turn limit on 88--97 equations, and Llama3.2-3B on 3--11. Qwen2-Math-1.5B does not support tool calling, so all of its requests are rejected (instead substituted with Qwen3.5-4B).

\textbf{Development Effort. }We count the lines of code for each method with \texttt{wc -l} (Table~\ref{tab:loc}). Function Calling needs 23 lines, a single call to a built-in function. The direct LLM integrations need 44 lines (Groq) and 40 (Ollama), since each is a single client-side HTTP call, to a billed API or to Ollama's local daemon. REST, SOAP and MCP require both a client and a server. REST takes 136 lines (75 client, 61 server) and SOAP 258 (87 client, 171 server), 1.9$\times$ REST's total and 2.8$\times$ its server code, driven by the hand-written WSDL contract and XML envelope and fault handling. MCP falls between the two at 165 lines (104 client, 61 server), as its tool schemas are generated from function signatures rather than written by hand. Relative to Function Calling, REST needs 5.9$\times$ more code, MCP 7.2$\times$ and SOAP 11.2$\times$. Building an agent on the same MCP server takes 259 lines (198 client), as much as SOAP and 6.5$\times$ the direct Ollama integration, so the agentic route is not cheaper to build either. These counts are a direct, measurable proxy for the engineering cost of the protocol rigor discussed in Section~\ref{sec:framework}.

\begin{table}[h]
\centering
\small
\begin{tabular}{lccc}
\hline
\textbf{Method} & \textbf{Client} & \textbf{Server} & \textbf{Total (LOC)} \\
\hline
Function Calling          & 23  & n/a & 23  \\
Ollama (local)            & 40  & n/a & 40  \\
Groq (remote API)         & 44  & n/a & 44  \\
REST                      & 75  & 61  & 136 \\
SOAP                      & 87  & 171 & 258 \\
MCP                       & 104 & 61  & 165 \\
LLM + MCP tools  & 198 & 61  & 259 \\
\hline
\end{tabular}
\caption{Lines of code required per method, excluding shared infrastructure (dataset loading, benchmark harness, CLI dispatch) common to all methods. The MCP server is shared by the MCP and LLM + MCP tools methods.}
\label{tab:loc}
\end{table}

\section{What to Consider in a Particular Situation}
\label{sec:framework}

Based on the experiments, we found that deterministic methods (Function Calling, REST, SOAP) achieve high dominance in all three metrics of accuracy, performance and cost optimization. LLMs answering directly are slower, less accurate and more compute-intensive regardless of model or temperature (generalist vs greedy decoding). Given deterministic tools, the best LLM agents approach deterministic accuracy, but only by delegating the arithmetic to those tools and at five to six orders of magnitude more latency. The practical difference is that deterministic methods require the task to already have an explicit implementation, as a library function or a deployed service, while LLMs can handle tasks specified in natural language. As stated in Section~\ref{sec:Introduction}, we compare architectural choices for functional invocation on arithmetic evaluation rather than on natural-language understanding. Based on these findings, we suggest the following:

\begin{itemize}
    \item \textbf{For accuracy-critical and high-stake tasks.} (e.g., financial calculations, safety-critical logic,), use deterministic methods (Function Calling, REST, SOAP or MCP). These were the only methods to reach 100\% accuracy.  


    \item \textbf{For cases with Strict protocol or formal contract requirements.} (e.g., regulated, multi-party or auditable exchange), prefer SOAP. Its WSDL discloses the full action space upfront and its fault format is fixed by the protocol, whereas REST leaves both to the implementation. MCP sits between them, declaring every tool and its argument schema upfront but returning errors as free-text messages, and is the natural choice when the callers are LLM agents that discover tools at runtime. Performance is comparable across all three (REST and MCP 326$\times$, SOAP 347$\times$ baseline), and code size grows with contract rigor (REST 136, MCP 165, SOAP 258 lines), so the choice rests on how much formal disclosure the setting requires.

    \item \textbf{For exploratory or prototyping work.} REST and LLM based settings work here. Direct LLM calls are the fastest to build (40--44 lines) and handle open-ended tasks, which makes them useful for exploration.  But development speed does not buy output trust (max. accuracy at 92.9\%). LLMs should not carry over into exploitation (prototyping and system build). Using REST as the deterministic alternative would be a good start (due to less requirements for WSDL definitions) once the task is understood would be a good start.
    
    \item \textbf{Latency-critical tasks:} avoid LLMs, which were 8{,}800$\times$ to 992{,}683$\times$ slower than the baseline. Among deterministic services, REST \& MCP are marginally faster than SOAP (326$\times$ vs 347$\times$), a negligible gap at this scalex.

    \item \textbf{Cost-constrained tasks:} the right choice depends on which kind of cost is being budgeted. Function Calling has effectively no marginal cost, and REST, SOAP and MCP add a bounded, predictable infrastructure cost (104--116$\times$ baseline processor time). A remote API's cost is metered by volume (39{,}180 tokens for 1{,}000 calls here), which is easy to forecast but recurs indefinitely. Local models replace that recurring bill with fixed hardware and energy costs, but consume orders of magnitude more compute (5{,}063$\times$ to 656{,}259$\times$ baseline processor time), and tool use multiplies it further.

\end{itemize}

Two further findings should inform this framework directly. First, \textbf{domain specialization does not guarantee deployability}: Qwen2-Math-1.5B, fine-tuned specifically for mathematical reasoning, scored 0.0\% in this benchmark due to inability to adhere to strict number-only output instruction (Table~\ref{tab:outcomes}), showcasing some pitfalls that can be faced from fine-tuning. Model selection process should verify instruction-following behavior directly rather than inferring from domain benchmarks. Phi4-mini-3.8B supports tool calling yet never called a tool, answering every equation itself, and the Llama models called tools but then answered in prose. Whether a model uses its tools must be measured, not assumed from its documentation. Second, \textbf{temperature is not a meaningful lever for this class of task}. Moving from $T=0$ to $T=0.7$ changed accuracy by at most 2.1 percentage points for direct invocation and 3.7 with tools. While the promise of 'determinism' is sought by many with $T=0$, all it does is Greedy decoding (always picks the single token with the highest predicted probability) which still is probabilistic rather than deterministic \cite{deterministic}.

\begin{tcolorbox}[colback=gray!5,colframe=gray!50!black,
title=Architectural Decision Framework (summary)]
\textbf{1. Is the procedure known and exact?} Use a deterministic method: Function Calling within a process, REST for a lightweight service, SOAP when a formal contract is required, and MCP when LLM agents will call the service. \\
\textbf{2. Is the procedure unknown and some error acceptable?} A direct LLM call is the fastest to build, but it should be replaced by a deterministic implementation once the task is understood. Where an LLM remains in the loop, it should call deterministic tools rather than compute the answer itself.
\end{tcolorbox}

\section{Discussion}
\label{sec:discussion}


We set out to show that established architectural alternatives remain viable and should be chosen by task type rather than by defaulting to a single approach. We quantified this via the proposal of the Architectural Decision Framework (Section~\ref{sec:framework}). The results support this premise. Deterministic methods reach 100\% accuracy at negligible cost, while every LLM answering directly falls short by wide and inconsistent margins at orders of magnitude more latency and compute, regardless of model, hosting or temperature. This points to a mismatch between autoregressive generation and exact computation, since the model must preserve operator precedence implicitly rather than by construction. This gap holds regardless of the model, host environment, or decoding temperature. LLM agents narrow the accuracy gap only after deterministic tool allocation but at cost of latency that is orders of magnitude higher than baseline. The overhead of REST, SOAP and MCP is acceptable for low-frequency services, although high-volume use would call for batching or persistent connections.



\textbf{Failure composition.} Table~\ref{tab:outcomes} separates wrong answers from format failures. Answering directly, the models mostly follow the output instruction but miscalculate. Wrong answers account for 540 of 554 failures for Llama3.2-1B at $T=0$, 249 of 257 for Llama3.2-3B, 177 of 182 for Phi4-mini-3.8B and 66 of 71 for Granite4-3B. Qwen2-Math-1.5B is the opposite extreme, with all 1{,}000 responses being format failures at both temperatures. Fine-tuning for math did not help it follow an instruction that general-purpose models of similar size handled. With tools, the failures shift from computation to interaction. The Llama models mostly called tools and then answered in prose (641--755 format failures). Phi4-mini-3.8B never used its tools, so its failures remain wrong answers (318--347).

\textbf{Temperature invariance. } At both greedy decoding ($T=0$) and commond default ($T=0.7$)changed accuracy by at most 2.1 points when answering directly (Llama3.2-1B, 44.6\% to 42.5\%) and 3.7 points with tools (Figure~\ref{fig:breakdown}b), with little effect on time or cost. Greedy decoding removes sampling variation but not model error, so $T=0$ makes an LLM's output more repeatable, not more correct. This agrees with prior findings that ``deterministic'' settings do not make LLMs deterministic~\cite{deterministic}.

These findings should be read within the scope of what we tested and the assumptions made in the process. Our benchmark is limited to algebraic expressions with a fixed prompt template, chosen for the ease of verifying answers and the need for structured understanding of operator precedence semantics, isolating the reasoning problem from more popular NLU-based math reasoning benchmarks \cite{hendrycks2021measuring, cobbe2021gsm8k}. We also evaluated only two temperature settings per model. Extending this comparison to tasks that mix computation with language understanding, or to cases with more complex semantics is a natural next step.


\textbf{Related Work.} Our work focuses on the appropriateness of architecture choices (such as REST, SOAP, MCP, LLMs) to get the same functionality. Others have started looking at appropriateness of LLMs for different settings. \citep{assadi2026embedder}, for example, compare embedding models against LLMs on text-based tasks. Their study finds near-identical aggregate quality ($77.6$ vs $77.2$) but at three orders of magnitude higher inference cost in dollars for the best LLM ($\$154$ vs $\$0.11$, a 1,431-fold difference). Embedding pipelines remain to stand as more efficient setting for classification, similarity and clustering based tasks. 

Together, these results indicate that the right architecture is a decision made under acceptable tradeoffs between accuracy, developer convenience, protocols adherence, performance and cost, and more (as mentioned in \ref{sec:framework}). We view this as the central implication of our arithmetic case study for the broader adoption of agentic systems in business automation.

\section{Conclusion}
\label{sec:conclusion}

In this paper, we propose an architectural decision framework (ADF) for selecting an invocation mechanism under trade-offs between accuracy, developer convenience, protocol adherence, performance and cost, grounded in a comparison of seven strategies on 1{,}000 arithmetic expressions: native function calls, REST, SOAP, MCP, LLMs answering directly (cloud and local), and LLM agents calling MCP tools. Deterministic methods scored 100\% accuracy at negligible cost, while LLMs answering directly reached at most 92.9\% at orders of magnitude more time and compute, regardless of model or temperature. LLM agents reached 99.9\% only by delegating the arithmetic to deterministic tools, at nearly 1{,}000{,}000$\times$ the latency of a function call, and weaker models performed worse with tools than without. Direct LLM calls were the easiest to integrate with least lines of code, but this ease did not translate into correctness, underscoring that no single axis of convenience, accuracy, or cost is sufficient on its own to justify an invocation choice. Extending the framework beyond this single task class, to richer semantics and multi-step workflows, is the natural next step. 


\bibliography{aaai2026}

\end{document}